\documentclass[letterpaper]{article} % DO NOT CHANGE THIS
\usepackage{aaai2026}  % DO NOT CHANGE THIS
\usepackage{times}  % DO NOT CHANGE THIS
\usepackage{helvet}  % DO NOT CHANGE THIS
\usepackage{courier}  % DO NOT CHANGE THIS
\usepackage[hyphens]{url}  % DO NOT CHANGE THIS
\usepackage{graphicx} % DO NOT CHANGE THIS
\usepackage{natbib}  % DO NOT CHANGE THIS AND DO NOT ADD ANY OPTIONS TO IT
\usepackage{caption} % DO NOT CHANGE THIS AND DO NOT ADD ANY OPTIONS TO IT
\usepackage{algorithm}
\usepackage{algorithmic}

\usepackage{newfloat}
\usepackage{listings}
\DeclareCaptionStyle{ruled}{labelfont=normalfont,labelsep=colon,strut=off} % DO NOT CHANGE THIS
\floatstyle{ruled}
\newfloat{listing}{tb}{lst}{}
\floatname{listing}{Listing}
\usepackage{makecell}
\usepackage{amsmath,amssymb,amsthm}
\usepackage{booktabs}
\usepackage{multirow}
\usepackage{subfigure}
\usepackage[nameinlink,capitalise,noabbrev]{cleveref}
\usepackage{tcolorbox}

\title{Rethinking the Sampling Criteria in Reinforcement Learning for LLM Reasoning: A Competence-Difficulty Alignment Perspective}
\title{Rethinking the Sampling Criteria in Reinforcement Learning for LLM Reasoning: A Competence-Difficulty Alignment Perspective}

\author{
    Deyang Kong\textsuperscript{\rm 1,2}\thanks{Equal contribution.},\;
    Qi Guo\textsuperscript{\rm 1,2}\footnotemark[1],\;
    Xiangyu Xi\textsuperscript{\rm 2}\thanks{Corresponding authors.},\\
    Wei Wang\textsuperscript{\rm 2},\;
    Jingang Wang\textsuperscript{\rm 2},\;
    Xunliang Cai\textsuperscript{\rm 2},\;
    Shikun Zhang\textsuperscript{\rm 1},\;
    Wei Ye\textsuperscript{\rm 1}\footnotemark[2]
}

\affiliations{
    \textsuperscript{\rm 1}National Engineering Research Center for Software Engineering, Peking University, Beijing, China\\
    \textsuperscript{\rm 2}Meituan Group, Beijing, China\\
    \{kong.deyang@foxmail.com, qguo@stu.pku.edu.cn, xixy10@foxmail.com, wye@pku.edu.cn\}
}

\usepackage{bibentry}
\begin{document}

\maketitle

\newcommand{\ourmethod}{\textbf{CDAS}}
\begin{abstract}

The low sampling efficiency during the rollout phase poses a significant challenge to scaling reinforcement learning for large language model reasoning.
Existing methods attempt to improve efficiency by scheduling problems based on problem difficulties. However, these approaches suffer from unstable and biased estimations of problem difficulty and fail to capture the alignment between model competence and problem difficulty in RL training, leading to suboptimal performance. 
To address these challenges, we introduce \textbf{C}ompetence-\textbf{D}ifficulty \textbf{A}lignment \textbf{S}ampling (\textbf{CDAS}). This approach allows for accurate and stable estimation of problem difficulties by aggregating historical performance discrepancies across problems. Subsequently, model competence is quantified to adaptively select problems whose difficulties align with the model's current competence using a fixed-point system.
Extensive experiments in mathematical RL training show that \textbf{CDAS} consistently outperforms strong baselines, achieving the highest average accuracy of 45.89\%. Furthermore, \textbf{CDAS} reduces the training step time overhead by 57.06\% compared to the widely-used Dynamic Sampling strategy, verifying the efficiency of \textbf{CDAS}. Additional experiments on different tasks, model architectures, and model sizes demonstrate the generalization capability of \textbf{CDAS}.
\end{abstract}

\begin{links}
    \link{Code}{https://github.com/DeyangKong/CDAS}
\end{links}


\section{Introduction}
Advanced large language models (LLMs) exemplified by DeepSeek-R1~\cite{guo2025deepseek} and OpenAI O1~\cite{openaio1} demonstrate remarkable performance in challenging tasks like mathematics. 
As a core technology in their reports, the Reinforcement Learning (RL) algorithm, such as Proximal Policy Optimization~\cite{schulman2017ppo} (PPO) and Group Relative Policy Optimization~\cite{shao2024deepseekmathgrpo} (GRPO), is employed to amplify the reasoning capabilities of the models.
It works by utilizing a verifier as the reward model to guide the generation of high-quality reasoning chains without the need for data annotation. 
Despite the promise, the RL training is costly and hard to scale, particularly due to its low sample efficiency during the rollout phase. Recent studies~\cite{yu2025dapo, zeng2025simplerl, bae2025online} indicate that sampling overly difficult problems often results in no correct chains, while sampling overly simple problems contributes little to model capabilities, leading to computational waste.
Consequently, a host of efforts are devoted to exploring sampling strategies for more efficient and stable RL training. 

Existing strategies draw inspiration from Curriculum Learning (CL)~\cite{bengio2009curriculum, narvekar2020curriculum}, scheduling data based on problem difficulty to enhance training stability and efficiency.
% (e.g., starting with simpler problems and gradually progressing to more challenging ones). The difficulty of an individual problem is typically modeled using a prior difficulty label or by utilizing the pass rate of models, with problems that exhibit higher pass rates being assigned lower difficulty levels.
% The primary of  lies in accurately modeling the difficulty of problems.
Curriculum Sampling Strategy~\cite{team2025kimi} relies on prior difficulty labels, which are excessively offline, neglecting the inherent capabilities of the model. Dynamic Sampling used by DAPO~\cite{yu2025dapo} demonstrates promising results by oversampling and filtering out problems with the pass rate equal to 1 and 0, which incurs substantial rollout overhead and compromises the training efficiency. 
Prioritized Sampling Strategy used by Kimi k1.5~\cite{team2025kimi} records the latest pass rate of each problem during training and adaptively assigns a higher sampling probability to those with lower pass rates. Overall, the pass rate has been widely adopted as a proxy for modeling problem difficulty.
\par
\begin{figure}[!hbt]
    \centering
  \includegraphics[width=0.85\linewidth]{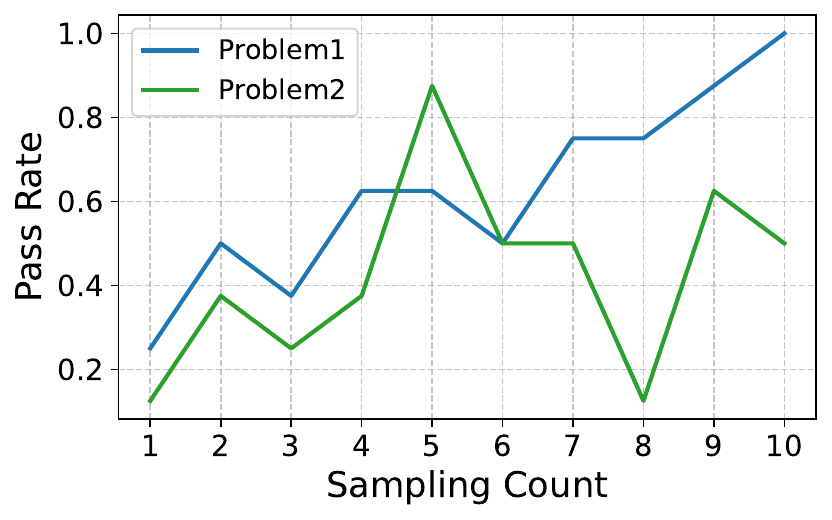}
    \caption{Pass rate variations of two problems in plain GRPO on Qwen2.5-7B using MATH dataset.}
    \label{fig:pass_rate_curve}
\end{figure}

However, these strategies tend to be suboptimal due to two main issues: 
% \par
(1) \textbf{Unstable and Biased Estimations of Problem Difficulty Using Single-Step Pass Rate}. 
As our experiment of training Qwen2.5-7B~\cite{yang2024qwen2p5} with MATH dataset~\cite{hendrycksmathdataset2021} shows (\cref{fig:pass_rate_curve}), the pass rate of individual problems exhibits considerable fluctuations throughout the training process, leading to unstable estimations of problem difficulty, consistent with several prior works~\cite{zheng2023secrets, peng2023stabilizing}. Moreover, focusing solely on the final pass rate can introduce a difficulty bias. 
% For instance, in \cref{fig:pass_rate_curve}, both problem 1 and problem 2 happen to achieve a final pass rate of 1, despite having distinct trajectories and difficulty levels. 
For instance, the pass rate of problem 2 at step 5 happens to surpass that of problem 1, despite their distinct trajectories and difficulty levels.
This demonstrates that the pass rate at a single step fails to capture the true complexity and learning dynamics associated with each problem.
(2) \textbf{Failing to Appropriately Capture the Alignment between Model Competence and Problem Difficulty}. A common strategy, such as in curriculum sampling and prioritized sampling, is to assign higher sampling probabilities to more difficult problems with lower pass rates.
A practical consequence is that overemphasizing difficult samples often leads to the selection of many zero-gradient problems with a pass rate of 0 in GRPO training, limiting the training efficiency~\cite{le2025no}.
However, a more appropriate approach, as advocated in the curriculum learning~\cite{bengio2009curriculum, narvekar2020curriculum}, is to prioritize problems that are more aligned with the current level of competency of the model for a more efficient and effective training.
\par

To address the issues above, we propose \textbf{C}ompetence-\textbf{D}ifficulty \textbf{A}lignment \textbf{S}ampling (\textbf{CDAS}) for RL training, dynamically sampling problems whose difficulty matches the model competence at the step level. 
% The core intuition behind \ourmethod \ is that instead of depending solely on the pass rate at a single step, an accumulative estimation that incorporates all historical information tends to provide a more stable measure of problem difficulty. Moreover, dynamically sampling problems whose difficulty matches the model capability facilitates more efficient training.
The core intuition behind \ourmethod \ is twofold: (1) instead of relying solely on the pass rate at a single step, an accumulative estimation that incorporates historical information tends to yield a more stable assessment of problem difficulty; and (2) explicitly modeling model competence to measure its alignment with problem difficulty, thereby enabling more effective sampling decisions. 
To realize these intuitions, we explicitly define two core concepts: \textbf{Model Competence} and \textbf{Problem Difficulty}. We then introduce \textbf{Competence-Difficulty Alignment} to quantify the gap between them, enabling us to sample problems that are optimally matched to the model's current learning stage. To ensure this difficulty estimation is stable and robust against single-step fluctuations, we aggregate the problem's \textbf{historical performance} over time. When integrated into the dynamics of RL training, this entire framework is formulated as a \textbf{fixed-point system}, which is theoretically guaranteed to converge, ensuring a stable and principled training process.
% Specifically, we model problem difficulty as the trajectory of performance discrepancy over training steps, where each point reflects the gap between the expected and actual pass rate. Then we use the centroid of this trajectory to provide a stable and accurate assessment of problem difficulty. Further, model competence is defined as the negative expected difficulty across all problems, and the absolute difference between model competence and a problem's difficulty is used to quantify their alignment.
% Considering the dynamics of RL training, the competence-difficulty alignment estimation above is further formulated as a difficulty-based fixed-point system, which can iteratively converge and ensures the stability of training with theoretical guarantees.
\par
\begin{figure}[t]
    \centering
    \subfigure[Average Accuracy]{
        \includegraphics[width=0.47\textwidth]{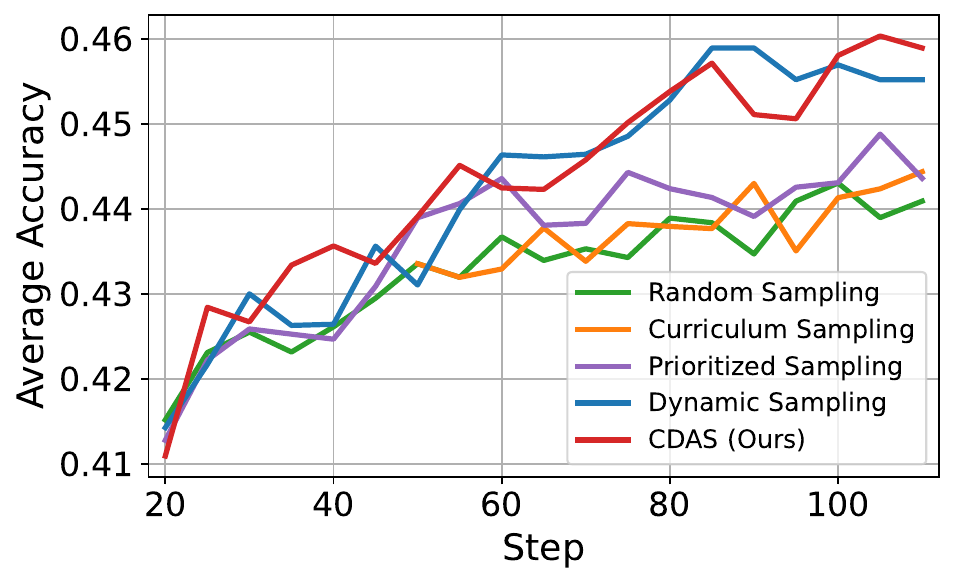}\label{fig:accuracy}
    }
    \subfigure[Total Training GPU Hours]{
        \includegraphics[width=0.47\textwidth]{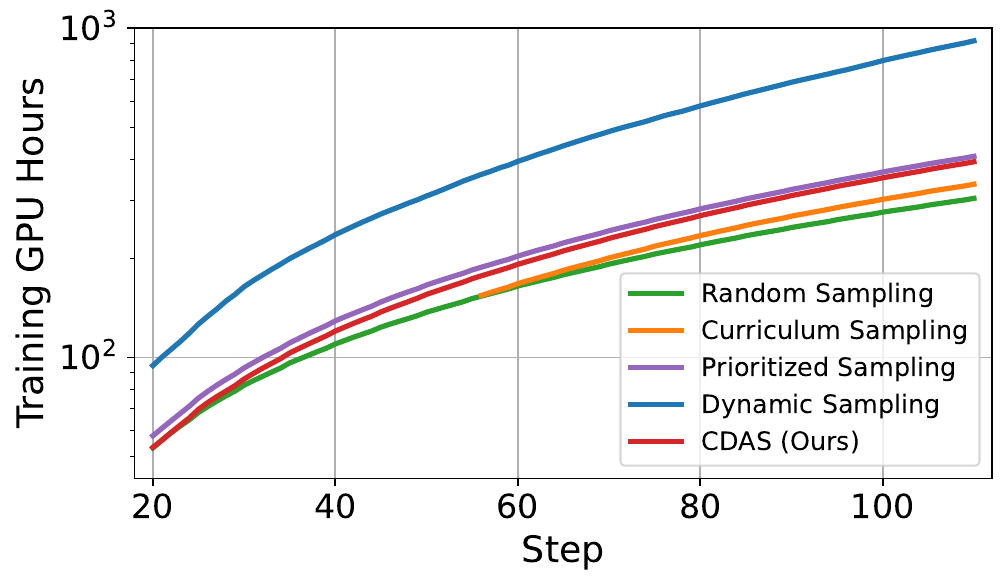}\label{fig:time}
    }
    \caption{Performance comparison of \ourmethod\ against baselines. \ourmethod\ achieves the best performance while demonstrating significant efficiency advantages compared to the strong Dynamic Sampling baseline.}
    \label{fig:main_results}
\end{figure}

To validate the effectiveness of \ourmethod, we conduct mathematical reinforcement learning on Qwen2.5-7B~\cite{yang2024qwen2p5} with GRPO, the most widely-used RL algorithm, and compare with extensive sampling strategies. The main findings are as follows: 
% 1. 主实验性能较好；比Dynamic Sampling效率占有
(1) \textbf{Superior Performance and Efficiency}: Results across six comprehensive mathematical reasoning benchmarks show that \ourmethod\ consistently outperforms powerful baselines, achieving the highest average accuracy of 45.89\%. Compared to Dynamic Sampling, a highly competitive baseline, \ourmethod\ reduces the training step time overhead by 57.06\% (see \cref{fig:time}), verifying the efficiency of \textbf{CDAS}. 
% 2. 采样效率高，验证了intuition
(2) \textbf{Efficient Sampling Mechanism}: In-depth analysis indicates that \ourmethod\  successfully samples more problems that align with the model’s competence, reducing the proportion of zero-gradient problems, significantly less than other sampling strategies.
% 3. generalization更好
(3) \textbf{Proven Generalization and Scalability}: Additional experiments on different tasks (i.e., code generation), model architectures (i.e., LLaMa), and model sizes (i.e., Qwen2.5-14B) demonstrate the generalization capability of \textbf{CDAS}.

The contributions of this paper can be summarized as follows:
\begin{itemize}
    \item We identify and analyze the limitations of existing sampling strategies from a new perspective, highlighting the importance of stable difficulty estimation and dynamic competence-difficulty alignment in RL training.
    \item We introduce \textbf{C}ompetence-\textbf{D}ifficulty \textbf{A}lignment \textbf{S}ampling (\ourmethod), adaptively selecting problems that match the model competence, which is grounded in a theoretically guaranteed fixed-point system. Extensive experiments validate its effectiveness and efficiency.
\end{itemize}
\par

\section{Preliminary}\label{preliminary}
\textbf{Group Relative Policy Optimization}
GRPO utilizes group-based advantage without a value model, thereby reducing computational overhead. 
Formally, given a problem $x$, the correct answer $y$, and a group of sampled responses $\{\hat{y_i}\}_{i=1}^{G}$ with their corresponding rewards $\{r_i\}_{i=1}^{G}$, GRPO calculates the advantage by normalizing the rewards within each group.
% 我们 study 中使用的是改进的 GRPO 算法
The original GRPO objective employs a sample-level loss calculation which potentially introduces length biases~\cite{yu2025dapo, liu2025understanding, chu2025gpg}, so we utilize a token-level policy gradient loss as our objective function:

\begin{equation}
\begin{split}
&\mathcal{J}_{\textrm{GRPO}}(\theta) = \mathbb{E}_{[x \sim \mathcal{D}, \{\hat{y_i}\}_{i=1}^G \sim \pi_{\textrm{old}}(\cdot|x)]} \\
&\quad \frac{1}{\sum_{i=1}^G |\hat{y_i}|} \sum_{i=1}^G \sum_{t=1}^{|\hat{y_i}|} \bigg(\min \Big(r_{i,t}(\theta) \hat{A}_{i,t}, \\
&\quad \text{clip} \Big(r_{i,t}(\theta), 1 - \epsilon, 1 + \epsilon\Big) \hat{A}_{i,t}\Big) - \beta \mathbb{D}_{\text{KL}}[\pi_\theta \parallel \pi_{\text{ref}}]\bigg),
\end{split}
\end{equation}

where
\begin{equation}
r_{i,t}(\theta) = \frac{\pi_\theta(\hat{y}_{i,t} \mid x, \hat{y}_{i,<t})}{\pi_{\theta_{\text{old}}}(\hat{y}_{i,t} \mid x, \hat{y}_{i,<t})},
\end{equation}\\
\begin{equation}
\hat{A}_{i,t} = \frac{r_i - \text{mean}(\{r_i\}_{i=1}^{G})}{\text{std}(\{r_i\}_{i=1}^{G})}.
\end{equation}
\textbf{Rule-Based Reward}
The use of reward model usually leads to reward hacking problem~\cite{gao2023scaling,weng2024rewardhack}, so we use a rule-based reward function. The reward is computed using the following rule:
\begin{equation}
r(y, \hat{y}) =
% r(y, \hat{y}) = 1 \ \text{if is\_equivalent}(y, \hat{y}) \ \text{else} \ 0
\begin{cases}
    1 &  \mathrm{is\_equivalent}(y, \hat{y}) \\
    0 & \mathrm{otherwise}
\end{cases}
\end{equation}

Notably, we do not employ a format reward. Prior research indicates that strict format constraints may limit the upper bound of model performance~\cite{zeng2025simplerl, singh2023beyond, wang2024planning}. Therefore, we directly use the final accuracy as the reward.

\section{Competence-Difficulty Alignment Sampling}
\label{das_method}

In this section, we introduce the \textbf{C}ompetence-\textbf{D}ifficulty \textbf{A}lignment \textbf{S}ampling (\ourmethod) framework. We first motivate the need for a dynamic and stable difficulty metric, then define model Competence, problem Difficulty, and their Alignment. Finally, we present the alignment-based sampling strategy and its convergence properties.

\subsection{Rethinking Problem Difficulty: The Need for Relativity and Stability}
The efficacy of advanced sampling strategies hinges on their ability to assess problem difficulty accurately. We argue that existing methods are limited by their failure to account for two aspects: the relativity and stability of difficulty.

\textbf{The Relativity of Difficulty.}
Static curriculum learning strategies rely on a fixed, absolute measure of difficulty (e.g., predefined labels). However, problem difficulty is not an intrinsic constant; it is inherently relative to the competence of the model attempting to solve it. 
A problem that is challenging early in training may become trivial as the model improves, so a meaningful difficulty metric must dynamically reflect the evolving relationship between the model’s capability and the problem’s complexity.

\begin{figure}[!h]
    \centering
    \includegraphics[width=0.85\linewidth]{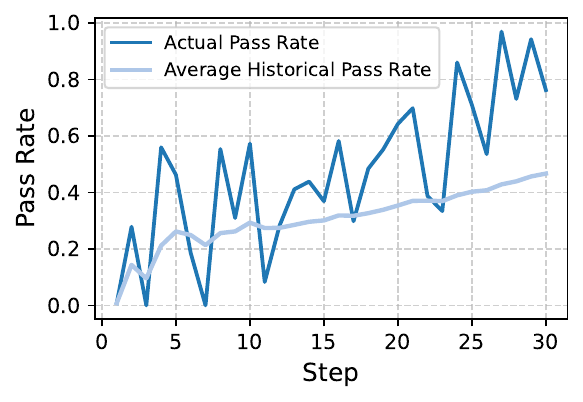}
    \caption{Pass Rate vs Step.}
    \label{fig:historical_pass_rate}
\end{figure}

\textbf{The Stability of Difficulty.}
Using the instantaneous pass rate as a difficulty proxy introduces high variance. As shown in \cref{fig:historical_pass_rate}, pass rates fluctuate significantly despite an overall upward trend, making single-step observations noisy and unreliable.
To achieve a more reliable estimation, it is necessary to aggregate the problem's \textbf{historical performance}, smoothing out short-term noise to better capture the true learning trend.

These limitations often lead to misalignment between model competence and problem difficulty, resulting in overly hard or easy samples that yield zero-gradient updates.
To address this, we propose a sampling method based on two core concepts: Model Competence and Problem Difficulty.

\subsection{Quantifying the Core Concepts: Competence, Difficulty, and Alignment}
To realize our vision of a dynamic and stable sampling method, we introduce three interconnected concepts to ensure a clear logical foundation.

\textbf{Step 1: Defining Model Competence ($C$)}
We define the model’s competence $C_n$
as a measure of its overall ability to solve problems at the training step 
$n$. Formally, competence is the negative mean of problem difficulty across the dataset $\mathcal{D}$:
\begin{equation}
   C_n =  -\mathbb{E}_{x \in \mathcal{D}}[D_n(x)]
   \label{eq:competence}
\end{equation}
where $D_n(x)$ is the difficulty of problem $x$, which we will precisely define next. The negative sign indicates that as the average problem difficulty decreases (i.e., the model solves them more effectively), the model's competence score $C_n$ correspondingly increases.

\textbf{Step 2: Defining Problem Difficulty ($D$)}
With the macro-level concept of competence established, we can now define the difficulty of an individual problem. We distinguish between a problem's volatile instantaneous difficulty ($d$) at a single step and its much more robust stable difficulty ($D$), which is informed by its history.

A problem's instantaneous difficulty, $d_n(x)$, at step $n$ is defined as the gap between the model's expected performance and its actual performance.
\begin{itemize}
    \item \textbf{Actual Performance} is measured by the model's pass rate on problem $x$ at step $n$, denoted as $s_n(x)$.
    
    \item \textbf{Expected Performance}, $\hat{P}_{M_n}(x)$, models the probability with which we anticipate the model will solve the problem. This expectation should naturally depend on the gap between the model's prior competence, $C_{n-1}$, and the problem's established difficulty, $D_{n-1}(x)$. A natural and widely used way to capture this dependency is through a Sigmoid transformation:
    \begin{equation}
        \hat{P}_{M_n}(x) = \sigma(C_{n-1} - D_{n-1}(x))
        \label{eq:exp_prob}
    \end{equation}
    The Sigmoid function smoothly maps the real-valued competence–difficulty gap to a probability in $[0, 1]$, a formulation also widely adopted in ability–difficulty modeling such as Item Response Theory and ELO rating systems~\cite{baker2001basics, lord2008statistical}. If competence far exceeds difficulty ($C \gg D$), the expected probability of success approaches 1. Conversely, if a problem is far too difficult for the model's competence ($C \ll D$), the probability approaches 0. When competence and difficulty are balanced ($C \approx D$), the outcome is most uncertain, with a success probability of 0.5.
\end{itemize}

The \textbf{instantaneous difficulty} $d_n(x)$ is thus the difference between these two quantities:
\begin{equation}
    d_n(x) = \underbrace{\sigma(C_{n-1} - D_{n-1}(x))}_{\text{Expected Performance}} - \underbrace{s_n(x)}_{\text{Actual Performance}}
\end{equation}
If $d_n(x) > 0$, the model underperformed relative to expectations, implying the problem was harder than anticipated at this step. If $d_n(x) < 0$, the model overperformed, suggesting the problem was easier.

Finally, to achieve the desired \textbf{stability}, we define a problem's stable difficulty, $D_{n}(x)$, as the centroid of its historical instantaneous difficulties, which smooths out noise over time:
\begin{equation}
    D_{n}(x) = \frac{1}{n} \sum_{k=1}^n{d_k}
    = \frac{n-1}{n} \cdot D_{n-1}(x) + \frac{1}{n} \cdot d_n(x)
\label{eq:difficulty_update}
\end{equation}

% Finally, to achieve the desired \textbf{stability}, we define a problem's stable difficulty, $D_{t_j}(x_j)$, as the centroid of its historical instantaneous difficulties, which smooths out noise over time.
% In practice, this is computed efficiently via an incremental update each time a problem $x_j$ is sampled (for the $t_j$-th time):
% \begin{equation}
%     D_{t_j}(x_j) = \frac{t_j-1}{t_j} \cdot D_{t_j-1}(x_j) + \frac{1}{t_j} \cdot d_n(x_j)
%     \label{eq:difficulty_update}
% \end{equation}

\textbf{Step 3: Quantifying Competence-Difficulty Alignment ($\mathcal{A}$)}
With definitions for both competence ($C$) and difficulty ($D$), we can now naturally quantify their alignment for any given problem. We define the alignment, $\mathcal{A}$, as the absolute difference between the model's competence and the problem's difficulty:
\begin{equation}
    \mathcal{A}(x, M_{n}) = |C_{n-1} - D_{n-1}(x)|
    \label{eq:alignment}
\end{equation}

This alignment value, $\mathcal{A}$, measures how closely a problem’s difficulty matches the model’s current learning frontier. A low $\mathcal{A}$ value indicates that a problem is well-matched to the model's present capabilities, making it a prime candidate for efficient and effective training.

\subsection{Alignment-based Sampling and Training}

\textbf{Symmetric Sampling Strategy}
A naive approach to leverage our alignment metric might be to simply sample problems with the lowest absolute value of $\mathcal{A}(x, M_n)$. However, this could cause the sampling distribution to become overly biased.
To foster a more balanced and robust training dynamic, we introduce a \textbf{Symmetric Sampling Strategy}. At each training step, we partition the entire problem set into two groups based on their relationship to the model's current competence:
\begin{itemize}
    \item \textbf{The Slightly Harder Group ($B^+$):} Problems for which the difficulty is greater than the model's competence ($D_{n-1}(x) > C_{n-1}$). These are problems that currently lie just beyond the model's ability.
    \item \textbf{The Slightly Easier Group ($B^-$):} Problems for which the difficulty is less than or equal to the model's competence ($D_{n-1}(x) \leq C_{n-1}$). These are problems that are within the model's current grasp.
\end{itemize}

We then construct the training batch $B$ by selecting the $|B|/2$ problems with the lowest alignment value $\mathcal{A}$ from each group, forming the final batch $B = B^- \cup B^+$. This balanced approach prevents the sampling process from being biased towards only one side. Further study of Symmetric Sampling Strategy is provided in Appendix~\ref{symmetric_sampling}.

\textbf{CDAS in Reinforcement Learning: A Stable Fixed-Point System}
We integrate \ourmethod\ framework into the GRPO training loop, as detailed in Algorithm~\ref{alg:das}.
Since \(|B|\) is usually much smaller than the size of the training set, performing a full update of problem difficulties at each step will lead to heavy computational overhead. Instead, for problem \(x_j\), we record the number of times it is sampled as \(t_j\) and update its difficulty only when it is sampled. Specifically, at the \(i\)-th step, the problem difficulty can be updated iteratively by
\begin{equation}
    D_{t_j}(x_j) = \frac{t_j-1}{t_j} \cdot D_{t_j-1}(x_j) + \frac{1}{t_j} \cdot d_n(x_j)
    \label{eq:difficulty_update}
\end{equation}

We set the initial difficulty $D_0(x)$ for all problems and the initial model competence $C_0$ to 0, signifying no prior system knowledge.

\begin{algorithm}[tb]
\small
\caption{Competence-Difficulty Alignment Sampling in GRPO}
\label{alg:das}
\textbf{Input}: Training set $\{(x_1, y_1), (x_2,y_2), \ldots, (x_n,y_n)\}$, Model $M_0$\\
\textbf{Parameter}: Total steps $K$, batch size $|B|$\\
\textbf{Output}: Updated model $M_K$
\begin{algorithmic}[1]
\STATE Initialize $C_{0} = 0$, $t_j=0$, $D_{t_j}(x_j) = 0$ for $j = 1,2,\ldots,N$
\FOR{$n = 1$ to $K$}
    \FOR{$j = 1$ to $N$}
        \STATE $\mathcal{A}(x_j, M_{n-1}) \gets |C_{n-1}-D_{t_j}(x)|$
    \ENDFOR
    \STATE Sample $B = B^- \cup B^+$ based on $\mathcal{A}$~~~~~~~//~GRPO Rollout
    \FOR{$(x_j, y_j)$ in $B$}
        \STATE Compute pass rate $s_n(x_j)$
        \STATE Compute rewards $r$ for each sampled response
        \STATE $t_j \gets t_j + 1$
        \STATE $D_{t_j}(x_j)\gets \frac{t_j-1}{t_j} \cdot D_{t_j-1}(x_j) + \frac{1}{t_j} \cdot (\sigma(C_{n-1}-D_{t_j-1}(x_j))- s_{n}(x_j))$
    \ENDFOR
    \STATE $C_{n} \gets -\mathbb{E}_x[D_{t_j}(x)]$
    \STATE Update policy model $M_{n+1}$ by maximizing GRPO objective
\ENDFOR
\STATE \textbf{return} $M_K$
\end{algorithmic}
\end{algorithm}

Notably, the dynamic and interdependent update process, where competence and difficulty co-evolve, can be rigorously analyzed as a \textbf{Fixed-Point System}. As training progresses and the model's parameters converge, we expect the system to reach a stable equilibrium state $(C^*, D^*, S^*)$, which satisfies the following conditions:
\begin{equation}
  \begin{cases}
  D^*(x) = \sigma(C^* - D^*(x)) - S^*(x), \ \ \forall x \in \mathcal{D} \\
  C^* = -\mathbb{E}_x[D^*(x)]
  \end{cases}
\end{equation}

Since the Sigmoid function is a contraction mapping, we can theoretically prove that this system is guaranteed to converge to a unique solution (proof provided in Appendix~\ref{proof}).
The physical meaning of this equilibrium state provides important insight into the model's final capabilities:
\begin{itemize}
    \item \textbf{Converged Pass Rate ($S^*(x)$):} Represents the stable, expected probability of solving problem $x$ once the system stabilizes.
    \item \textbf{Converged Difficulty ($D^*(x)$):} Measures the gap between problem complexity and model capability. Positive values denote unsolved challenges; negative values denote mastery.
    \item \textbf{Converged Competence ($C^*$):} Represents the model's final, overall capability ceiling on the given task, averaged across all problems.
\end{itemize}

\section{Experiments}

\begin{table*}[!ht]
\centering
\small
\begin{tabular}{cl|cccccccc}
\toprule
\textbf{Steps} & \textbf{Method} & \makecell[c]{\textbf{AIME24}\\(Avg@32)} & \makecell[c]{\textbf{AIME25}\\(Avg@32)} & \makecell[c]{\textbf{MATH}\\\textbf{500}} & \makecell[c]{\textbf{Minerva}\\\textbf{Math}} & \makecell[c]{\textbf{Olympiad}\\\textbf{Bench}} & \textbf{GSM8K} & \textbf{Avg.} \\
\midrule
0 & - & 0.06 & 0.00 & 47.00 & 17.65 & 14.52 & 81.50 & 24.38 \\
\midrule
\multirow{5}{*}{55}
 & Random Sampling & 10.00 & 7.29 & 74.20 & 36.76 & 39.26 & 91.66 & 43.20 \\
 & Curriculum Sampling & 10.00 & 7.29 & 74.20 & 36.76 & 39.26 & 91.66 & 43.20 \\
 & Prioritized Sampling & \underline{11.77} & 7.08 & 74.00 & \textbf{39.34} & \underline{40.15} & \underline{92.04} & \underline{44.06} \\
 & Dynamic Sampling & \textbf{12.19} & \textbf{7.92} & \underline{75.00} & \underline{38.97} & 38.52 & 91.36 & 43.99 \\
 & CDAS (Ours) & 11.56 & \underline{7.71} & \textbf{76.20} & \underline{38.97} & \textbf{40.44} & \textbf{92.19} & \textbf{44.51} \\
\midrule
\multirow{5}{*}{110}
 & Random Sampling & 12.29 & 6.98 & 75.20 & 37.13 & \underline{40.00} & \textbf{92.95} & 44.09 \\
 & Curriculum Sampling & 12.71 & 7.19 & \underline{76.00} & 38.60 & 39.56 & \underline{92.57} & 44.44 \\
 & Prioritized Sampling & \textbf{15.10} & 9.27 & 75.00 & 37.50 & 39.56 & 91.51 & 44.66 \\
 & Dynamic Sampling & \textbf{15.10} & \underline{9.58} & \textbf{77.20} & \underline{39.34} & 39.56 & 92.34 & \underline{45.52} \\
 & CDAS (Ours) & \underline{14.90} & \textbf{11.77} & 75.40 & \textbf{40.44} & \textbf{40.89} & 91.96 & \textbf{45.89} \\
\bottomrule
\end{tabular}
\caption{Performance comparison across different sampling methods on various math benchmarks. Metrics are Avg@32 for AIME and standard accuracy for others. We present the best results in \textbf{bold} and the second with 
\underline{underline}.}
\label{tab:main_res}
\end{table*}

\subsection{Experimental Setup}\label{setup}
% \textbf{Dataset} We use MATH dataset~\cite{hendrycksmathdataset2021} as our training dataset. Following ,we reserve the MATH500 subset as the test set, and combine remaining 4500 test problems with 7500 training problems as the training dataset.
\textbf{Dataset} Following~\citet{zeng2025simplerl}, we utilize the MATH dataset~\cite{hendrycksmathdataset2021} for RL training, including 7,500 training samples and 4,500 test samples. We reserve its MATH500 subset as the validation set for RL training.\\
\textbf{Baselines} We compare \ourmethod \ against a range of powerful baselines: (1) \textbf{Random Sampling}, which directly samples the problem randomly from the training dataset and can be viewed as plain GRPO training. (2) \textbf{Curriculum Sampling}~\cite{team2025kimi}, which uniformly samples easy problems first, then turns to harder ones.
Here, we use the difficulty level tags included in the MATH dataset, ranging from 1 to 5, and conduct training from the middle checkpoint in Random Sampling on problems with difficulty level \(\geq 4\). (3) \textbf{Prioritized Sampling}, which is adopted in Kimi k1.5~\cite{team2025kimi} that tracks the pass rate $s$ of each problem and then samples problems based on $1-s$.(4) \textbf{Dynamic Sampling}, adopted in DAPO~\cite{yu2025dapo}, which over-samples and filters out problems with the pass rate equal to 1 or 0.

\textbf{Training Details}
We experiment with GRPO, the most widely-used algorithm for LLM RL training, on Qwen2.5-7B~\cite{yang2024qwen2p5} using the veRL framework~\cite{sheng2024hybridflowverl}. Following the training recipe in SimpleRL-Zoo~\cite{zeng2025simplerl}, we set the batch size \(|B| = 1024\), generating \(8\) rollouts for each problem with temperature \(1.0\) and maximum response tokens \(4096\). Refer to Appendix~\ref{othertrainingdeatils} for more training details.

Considering the stability of our iterative system, we adopt a \textbf{warm-up strategy} in \ourmethod. For the duration of the first training epoch, we sample problems uniformly at random. This ensures every problem is sampled at least once, allowing us to establish a robust initial value for each problem's difficulty. By doing so, we directly mitigate the risk of amplifying dataset biases before the main adaptive sampling phase begins. Further analysis is available in Appendix~\ref{warmup_phase}.

\textbf{Evaluation} For a comprehensive evaluation, we select 6  mathematical reasoning tasks, including AIME 24\&25, MATH500~\cite{hendrycksmathdataset2021}, Minerva Math~\cite{lewkowycz2022solvingminerva}, Olympiad Bench~\cite{he2024olympiadbench} and GSM8K~\cite{cobbe2021traininggsm8k}.
Since the number of problems in AIME24 and AIME25 is relatively small, we report the Avg@32 metric to reduce randomness. Standard accuracy is reported for the remaining tasks. More details are shown in Appendix~\ref{benckmarks}.

\subsection{Main Results}

\begin{figure*}[!ht]
    \centering
    \subfigure[Random Sampling and Curriculum Sampling]{
        \includegraphics[width=0.45\textwidth]{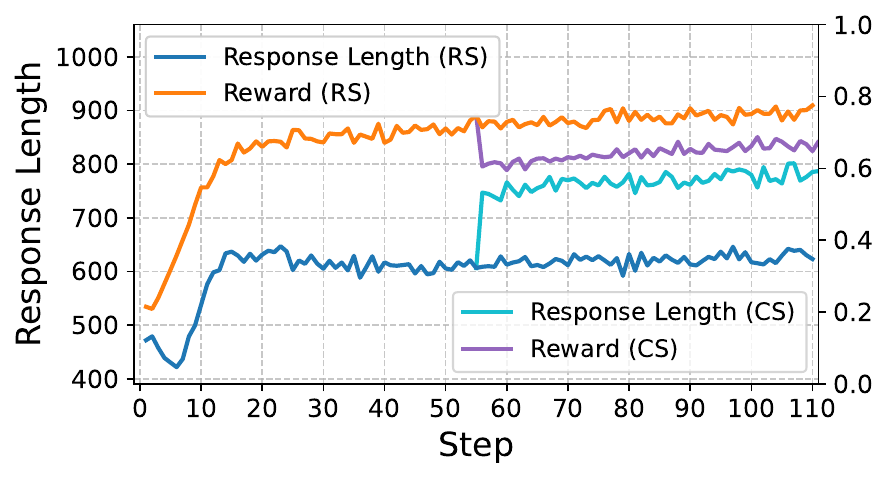}
    }
    \subfigure[Prioritized Sampling]{
        \includegraphics[width=0.45\textwidth]{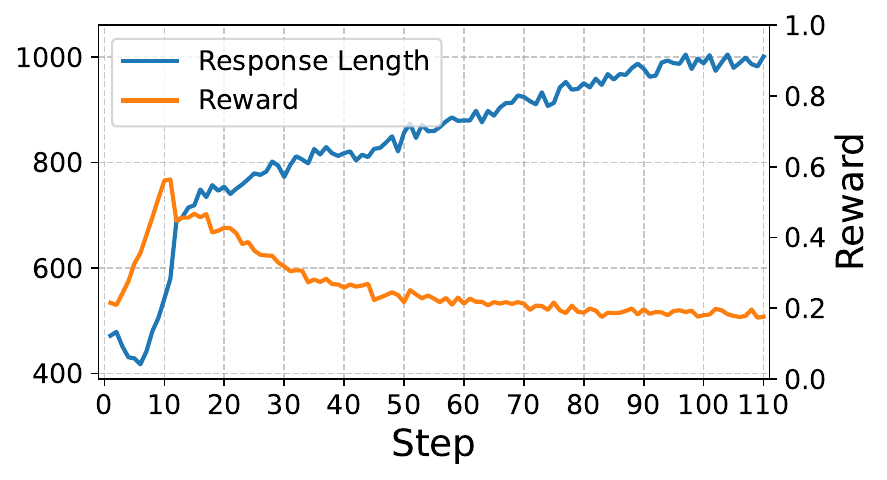}
    }
    \subfigure[Dynamic Sampling]{
        \includegraphics[width=0.45\textwidth]{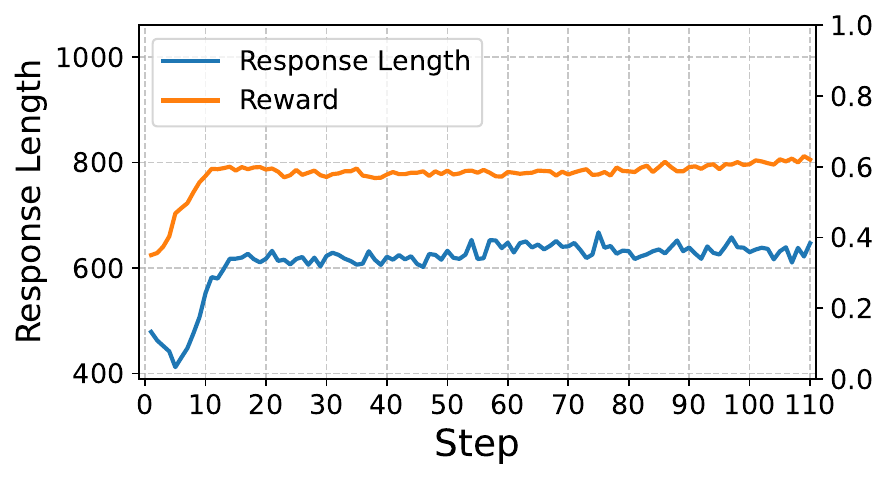}
    }
    \subfigure[CDAS]{
        \includegraphics[width=0.45\textwidth]{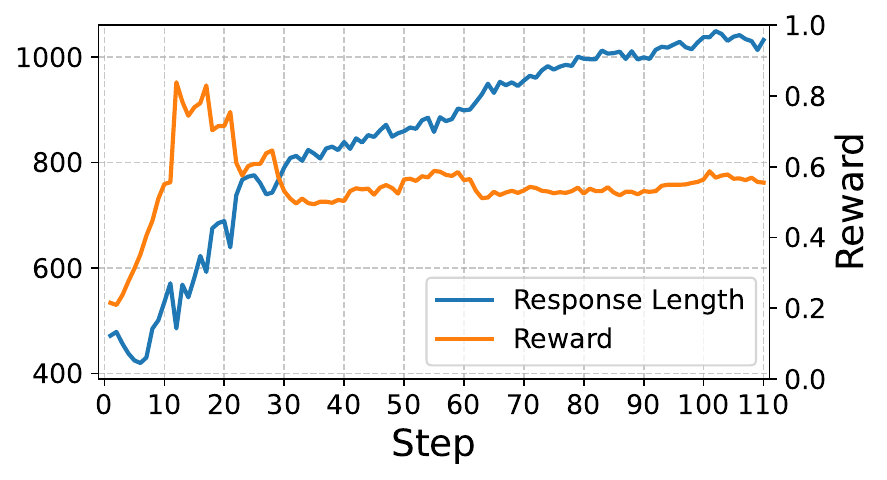}
    }
    \caption{Training curves of different sampling strategies.}
    \label{fig:train}
\end{figure*}

The main results are summarized in \cref{tab:main_res}, with training curves shown in \cref{fig:train} and performance comparison curves shown in \cref{fig:main_results}. 
Besides the final checkpoint, we also report the intermediate checkpoint at the 55-th step. From the results, we have the following findings:
\par
\textbf{CDAS outperforms plain GRPO and other baselines.} As shown in \cref{tab:main_res}, CDAS achieves the best average accuracy at both the 55-th step (44.51\%) and the 110-th step (45.89\%) checkpoints. Meanwhile, we find that the improvements offered by Curriculum Sampling (+0.35\% at the 110-th step) and Prioritized Sampling (+0.57\% at the 110-th step) over the Random Sampling baseline are limited. This indicates that relying solely on prior difficulty labels or single-step pass rates tends to be suboptimal.

\par
\textbf{CDAS demonstrates a substantial efficiency advantage.} In terms of sample efficiency, \ourmethod\ achieves a performance of 44.51\% at just 55 training steps, surpassing the final performance of Random Sampling (44.09\% at 110 steps), reaching comparable capability in half the training time. Moreover, as shown in \cref{fig:time}, the training time of \ourmethod\ is comparable to most standard baselines. This stands in stark contrast to Dynamic Sampling, which incurs significantly higher computational costs (2.33× slower).
% when compared to the competitive Dynamic Sampling, \ourmethod\ achieves superior final performance while being far more computationally efficient. 
% As shown in \cref{tab:main_res}, Dynamic Sampling's per-step time is \textbf{2.33X} that of \ourmethod\ (3745s vs. 1608s), highlighting the immense overhead.
% \par
% \textbf{CDAS consistently enhances the performance on more challenging benchmarks}. 
% On difficult benchmarks like AIME24/25, Minerva Math and Olympiad Bench, \ourmethod \ demonstrates remarkable improvements. For instance, \ourmethod \ achieves 11.77\% accuracy on AIME25 and 40.89\% on Olympiad Bench, notably higher than baselines. 
% However, on simpler benchmarks like MATH and GSM8K, the accuracy does not continue to improve after 55 steps in subsequent training. 
% Taking GSM8K as an example \cref{gsm8k}, \ourmethod \ achieves the highest accuracy of \(93.32\%\) at the \(80\)-th step, then oscillates.
\par
% \textbf{The reward curve in \ourmethod \ initially grows and then converges to a median value}. From \cref{fig:train}, after 30 steps, the amplitude of reward narrows to within the range of 0.4 to 0.6 and then converges around 
% 0.5. This indicates that \ourmethod \ effectively selects problems with difficulties appropriate for the model. On the other hand, since Prioritized Sampling tends to select harder problems, its reward demonstrates a continuous downward trend.
\textbf{The reward curve in \ourmethod \ demonstrates a dynamic process.}
As shown in \cref{fig:train}, the reward curve for \ourmethod\ exhibits four distinct phases:
(1) \textbf{Warm-up Phase:} During the first epoch, the curve mirrors that of Random Sampling, as problems are sampled uniformly to establish an initial, unbiased difficulty assessment.
(2) \textbf{Rapid Ascent Phase:} Immediately following the warm-up, the reward sharply increases to over 0.8. This is because \ourmethod\ identifies and prioritizes the problems that have become easy for the model.
(3) \textbf{Adaptive Descent Phase:} As the model's competence grows, \ourmethod\ intentionally transitions to sampling more challenging problems that better match the model's new capabilities. This strategic shift causes the average reward to decrease.
(4) \textbf{Convergence Phase:} The curve finally stabilizes around a median reward of approximately 0.55, which signifies that \ourmethod\ continuously presents the model with problems within its competence.

% (1) initial improvement as the model efficiently learns from easier problems, (2) a brief drop corresponding to the sampler's intentional transition to harder problems, and (3) eventual stabilization around a median reward of 0.5. This indicates that \ourmethod\ successfully calibrates the problem difficulty to match the model's evolving competence. On the other hand, since Prioritized Sampling tends to select harder problems, its reward demonstrates a continuous downward trend. 

\par
% \textbf{An increased response length is not necessary for better performance}. Although we observed a rise in response length in \ourmethod\ and Curriculum Sampling, the response length in Dynamic Sampling stabilized after about 15 steps, yet the average accuracy on benchmarks continued to grow.

\section{Analysis and Discussion}
In this section, we conduct a comprehensive analysis and discussion of \ourmethod. We explore the statistical properties of \ourmethod \ in terms of sample utility, investigate its advantages over pass rate-based methods, and validate its generalization to code generation tasks along with different architectures and model sizes.
\subsection{Utility of the Sampled Problems}
\begin{figure}[!h]
\centering
\includegraphics[width=0.98\linewidth]{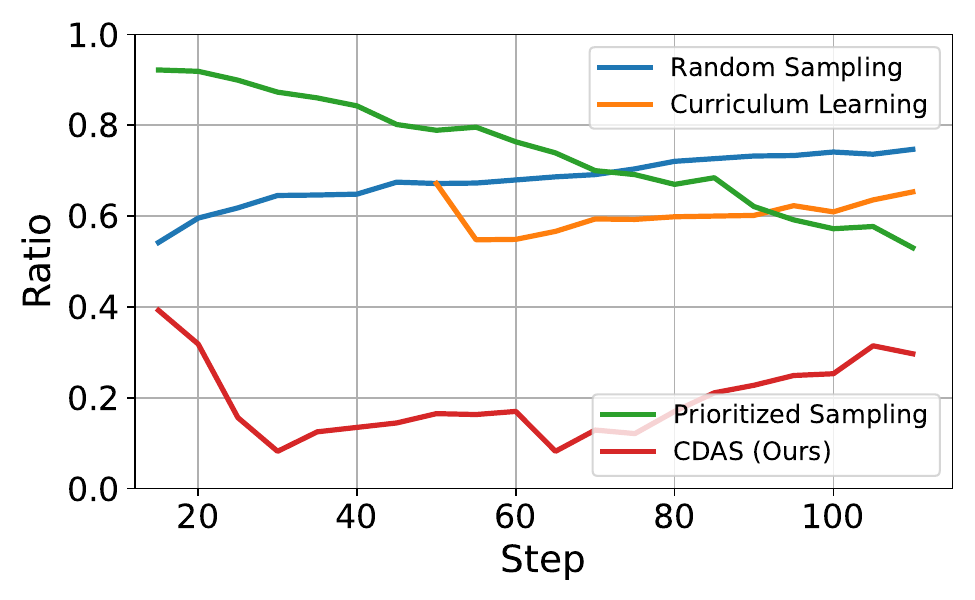}
\caption{The proportion of zero-gradient problems in the sampled batch.}
\label{fig:utility}
\end{figure}
From the perspective of the optimization objective of GRPO, the superior performance of Dynamic Sampling can be attributed to its filtering out of samples that do not contribute to model gradients~\cite{yu2025dapo} (i.e., those with a pass rate of \(0\) or \(1\)). Although \ourmethod \ does not explicitly constrain the pass rate in problem selection, its alignment-based symmetric sampling inherently mitigates the issue of oversampling the zero-gradient problems to some extent. As illustrated in \cref{fig:utility}, the proportion of such zero-gradient problems within the batches sampled by \ourmethod \ is consistently lower than that of the other baselines, proving that \ourmethod \ can effectively improve the utility of sampled problems. We also observe that the proportion of zero-gradient problems in \ourmethod \ exhibits a rapid decline during the early stages of training, followed by a slight increase in the later stages. The sharp decrease in the initial phase can be attributed to the swift correction of problem difficulty from its initial values. The modest rise in the later phase is mainly due to the increasing proportion of zero-gradient problems in the whole MATH training set, leading to more problems with a pass rate of 0 sampled in the batch \(B^+\).

\subsection{Problem Difficulty vs. Pass Rate}
\begin{figure}
    \centering
    \includegraphics[width=\linewidth]{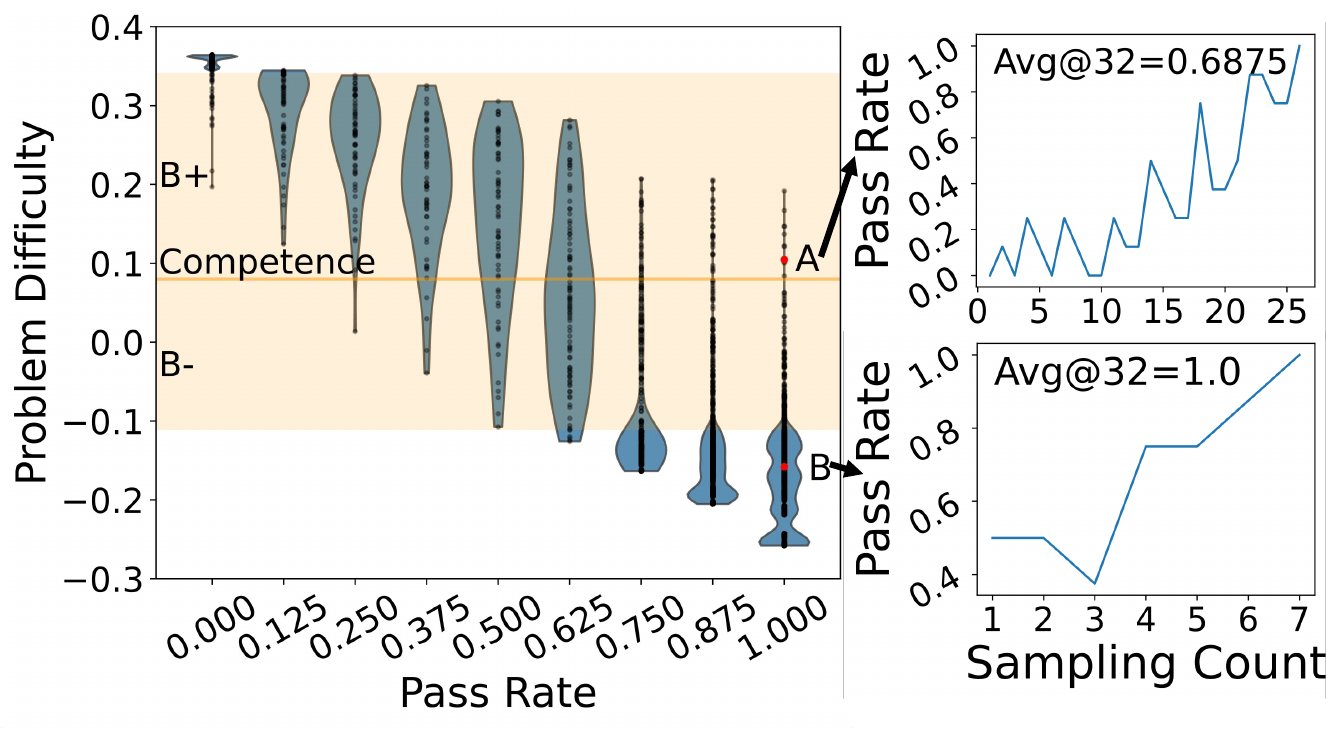}
    \caption{Problem difficulty vs. pass rate in \ourmethod.}
    \label{fig:case_study}
\end{figure}

Since the problem difficulty in \ourmethod\ derived from the pass rates, we explore the relationship between them. As shown in \cref{fig:case_study}, problem difficulty and pass rate exhibit an overall negative correlation where problems with lower difficulty tend to have higher pass rates. 
Interestingly, we find that even among problems with the same pass rate, there can still be considerable differences in their estimated difficulties. To further investigate this phenomenon, we randomly selected two problems with a pass rate of 1 at the final sampling step. 
As shown in \cref{fig:case_study}, problem A required 25 samplings to reach a pass rate of 1, whereas problem B achieved a pass rate of 1 after only 6 samplings.
Despite both having the same final pass rate in rollout, Problem A is noticeably more difficult than Problem B, as indicated by its average accuracy of 32 inferences being 0.6875, which is much lower than that of Problem B (Avg@32 = 1.0). This validates that \ourmethod, by leveraging historical information, provides a more accurate and robust measure of problem difficulty.
% there will be differences in difficulty even between problems with the same pass rate. \cref{fig:case_study} plots the relationship between problem difficulty and pass rate at the last step. As the pass rate decreases, the overall difficulty of the samples shows an upward trend, with the difficulty still ranges from \(-0.3\) to \(0.2\) from problems whose pass rate \(=1\). We randomly selected two samples with different difficulty and plot their pass rates against sampling times. 
% We find that the pass rate rapidly reached \(1\) after only \(6\) samplings for the less difficult problem , while the pass rate of the more difficult problems gradually converged to \(1\) as the sampling times increased to \(25\). 
% Such observation is consistent with the modeling of difficulty from a trajectory perspective in \cref{modeling_diff}.  

\subsection{Generalization}
We further investigate the generalization capabilities of \ourmethod\ across different tasks, model architectures, and model sizes.
We compare our method against two representative sampling strategies: \textbf{Random Sampling (RS)}, the most widely used baseline, and \textbf{Dynamic Sampling (DS)}, a strong but computationally expensive performance baseline.

\textbf{Generalization to Code Generation Tasks} 
We apply \ourmethod\ to the code generation task. Specifically, we aggregate 10k open-source competitive programming problems from Apps~\cite{hendrycks2021apps}, Taco~\cite{li2023taco}, and CodeContests~\cite{li2022codecontest}, and perform GRPO training on Qwen2.5-Coder-7B~\cite{hui2024qwen2d5coder} for 100 steps. As shown in~\cref{tab:lcb}, CDAS achieves \textbf{18.24\%} Pass@1 accuracy on LiveCodeBench v5~\cite{jain2025livecodebench}, outperforming RS by \textbf{+3.00\%} and nearly matching DS. This verifies \textbf{CDAS}'s strong generalization to different tasks. We leave the comparison of other tasks for future work.

\begin{table}[h]
\centering
\begin{tabular}{cc}
\toprule
\textbf{Method} & \textbf{Pass@1 Acc.} \\
\midrule
RS & 15.24 \\
DS & \textbf{18.32} \\
CDAS    & \underline{18.24} \\
\bottomrule
\end{tabular}
\caption{Accuracy comparison on LiveCodeBench v5.}
\label{tab:lcb}
\end{table}

\textbf{Generalization to Different Architectures and Model Sizes}
To further examine the generalization capability of \ourmethod\ across architectures and scales, we conduct additional RL experiments on Qwen2.5-14B~\cite{yang2024qwen2p5} and OctoThinker-8B-Hybrid~\cite{wang2025octothinker}.
Both models are trained on the MATH dataset used in the main experiments.
For Qwen2.5-14B, we train for 200 steps with a batch size of 256, while OctoThinker-8B-Hybrid adopts the same hyperparameter configuration as the main setup.
As shown in~\cref{tab:cdas_comparison}, \ourmethod\ consistently outperforms the RS baseline and achieves performance that matches or exceeds the strong DS baseline on both models, while maintaining comparable training efficiency to RS.
These results demonstrate the robust generalization of CDAS to models of larger scale and different architectural designs.

% We further validate the generalization ability of \ourmethod\ with different architectures and sizes. Specifically, we conduct mathematical RL training on Qwen2.5-14B~\cite{yang2024qwen2p5} and OctoThinker-8B-Hybrid~\cite{wang2025octothinker}. 
% As shown in~\cref{tab:cdas_comparison}, \ourmethod\ consistently outperforms RS and surpasses or matches the strong DS baseline across both models, while preserving training efficiency comparable to RS. This validates \textbf{CDAS}'s robust generalization to larger models and different architectures.

% In addition to the strong performance observed on Qwen2.5-7B, we further validate the effectiveness of \ourmethod \ on larger LLMs. Specifically, we conduct training on Qwen2.5-14B\cite{yang2024qwen2p5} with a batch size of 256 for 200 steps, matching the computational budget of our main experiments. The average accuracy of \ourmethod \ and Random Sampling is reported in \cref{fig:14b}. We find that \ourmethod \ achieves substantial
% improvements over the Random Sampling baseline by \(+1.47\%\), which is even greater than the improvement observed on the 7B model \(+1.27\%\), showing effectiveness on larger models.

\begin{table}[!ht]
\centering
\begin{tabular}{ccc}
\toprule
\textbf{Model} & \textbf{Method} & \textbf{Average Acc.} \\
\midrule
\multirow{3}{*}{Qwen2.5-14B} & RS & 46.28 \\
 & DS & \underline{46.62} \\
 & CDAS & \textbf{47.22} \\
\midrule
\multirow{3}{*}{OctoThinker-8B} & RS & 35.30 \\
 & DS & \underline{36.47} \\
 & CDAS & \textbf{36.52} \\
\bottomrule
\end{tabular}
\caption{Generalization Performance Across Different Architectures and Model Sizes.}
\label{tab:cdas_comparison}
\end{table}

\section{Conclusion}
We present \textbf{C}ompetence-\textbf{D}ifficulty \textbf{A}lignment \textbf{S}ampling (\textbf{CDAS}), a novel sampling strategy for RL training in LLM reasoning. \ourmethod \ addresses the limitations of existing methods by modeling problem difficulty as a trajectory of performance discrepancies to provide more stable estimations and explicitly aligning it with model competence at each training step throughout a fixed-point system. Extensive experiments on mathematical reasoning benchmarks demonstrate the superiority of \ourmethod\ in both accuracy and efficiency to powerful baselines. Our results highlight the importance of dynamically matching problem difficulty to model competence for efficient RL training.

\bibliography{aaai2026}

\newpage
\appendix
\onecolumn

\section{Related Work}
% \subsection{Reinforcement Learning with LLMs}
\textbf{RL for LLMs reasoning}.
Reinforcement learning (RL) has been widely adopted to enhance the reasoning abilities of LLMs, especially in mathematics and programming tasks~\cite{he2024olympiadbench, trinh2024solving, jain2025livecodebench, penedo2025codeforces}.
% 现在流行的 RL 可以划分为两类。首先是 actor-critic-base methods，这类方法需要 value-model，计算成本较高，如 ppo，还有字节最新的 vapo
Actor-critic-based methods, such as Proximal Policy Optimization (PPO)~\cite{schulman2017ppo}, utilize a value model to estimate the value function, guiding the policy updates.
% Value Augmented Proximal Policy Optimization (VAPO)~\cite{yuan2025vapo} incorporates Value-Pretraining and Decoupled-GAE to mitigate value model bias over long sequences.
% 其次是 REINFORCE-based methods，不需要 valued-model，计算成本较低，例如 grpo 及其变体（REINFORCE++, dapo等）
On the other hand, REINFORCE-based methods rely on policy gradients without a value model. Group Relative Policy Optimization (GRPO)~\cite{shao2024deepseekmathgrpo} normalizes rewards within a group of generated outputs, eliminating the need for a separate value model. REINFORCE++~\cite{hu2025reinforce++} enhances the classical REINFORCE algorithm~\cite{williams1992simple} by incorporating optimization techniques from PPO. Dynamic Sampling Policy Optimization (DAPO)~\cite{yu2025dapo} introduces several optimizations to enhance training efficiency and stability in long-CoT reasoning tasks.\\
% \subsection{Sampling Strategies for RL training}
\textbf{Sampling Strategies for RL training}.
Effective sampling strategies are crucial for efficient RL training with LLMs.
% 粗粒度的课程采样
Coarse-grained curriculum learning~\cite{team2025kimi,xie2025logic} gradually increases problem difficulty based on predefined labels.
LIMR~\cite{li2025limr} introduces Learning Impact Measurement (LIM) to select problems that align with the model's learning trajectory.
% kimi, dapo, srpo 的工作
Prioritized Sampling~\cite{team2025kimi} tracks the pass rate for each problem and samples problems in proportion to their failure rates. This approach directs the model's focus toward more challenging problems.
Dynamic Sampling~\cite{yu2025dapo} continues to sample problems within a batch until their pass rates are neither 0 nor 1.

\section{Convergence Analysis}
\label{proof}
Recall the following coupled fixed-point system:
$$
  \begin{cases}
    D^*(x) = \sigma(C^* - D^*(x)) - S^*(x), \quad x \in \mathbb{X} \\
    C^* = -\mathbb{E}_x[D^*(x)]
  \end{cases}
$$
where \(\sigma(z) = \frac{1}{1+e^{-z}}\) is the sigmoid function, and \(S^*(x)\) is a constant with \(x\).

Let \(\mathbf{D} = (D(x_1), D(x_2), ..., D(x_N))^\top\) denote the vector of problem difficulties, and \(C\) denote the model competence. The system can be written as a joint mapping \(\mathcal{F}\) in \((N+1)\)-dimensional space:
\[
\begin{cases}
D_{n+1}(x) = \sigma(C_n - D_n(x)) - S^*(x) \\
C_{n+1} = -\frac{1}{N}\sum_{i=1}^N D_{n+1}(x_i)
\end{cases}
\]
or equivalently, \((\mathbf{D}_{n+1}, C_{n+1}) = \mathcal{F}(\mathbf{D}_n, C_n)\).

The sigmoid function \(\sigma(z)\) is Lipschitz continuous\cite{rudin2021principles} with constant \(L_\sigma = \frac{1}{4}\). For any \(D(x), D'(x)\) and \(C, C'\), we have:
\[
|\sigma(C - D(x)) - \sigma(C' - D'(x))| \leq \frac{1}{4}\left( |C - C'| + |D(x) - D'(x)| \right)
\]

For two states \((\mathbf{D}, C)\) and \((\mathbf{D}', C')\), define the distance as
\[
\|(\mathbf{D}, C) - (\mathbf{D}', C')\| = \max\left\{ \max_x|D(x) - D'(x)|, |C - C'| \right\}
\]
Then,
\[
|D_{n+1}(x) - D'_{n+1}(x)| \leq \frac{1}{4} \left( |C_n - C'_n| + |D_n(x) - D'_n(x)| \right)
\]
Taking the maximum over all \(x\), we obtain
\[
\max_x|D_{n+1}(x) - D'_{n+1}(x)| \leq \frac{1}{4} \left( |C_n - C'_n| + \max_x|D_n(x) - D'_n(x)| \right) \leq \frac{1}{2}\delta_n
\]
where \(\delta_n = \max\left\{ \max_x|D_n(x) - D'_n(x)|, |C_n - C'_n| \right\}\).

For the update of \(C\), we have
\[
|C_{n+1} - C'_{n+1}| = \left| -\frac{1}{N}\sum_x D_{n+1}(x) + \frac{1}{N}\sum_x D'_{n+1}(x) \right| \leq \max_x|D_{n+1}(x) - D'_{n+1}(x)|
\]
Therefore, the joint mapping satisfies.
\[
\|(\mathbf{D}_{n+1}, C_{n+1}) - (\mathbf{D}'_{n+1}, C'_{n+1})\| \leq \frac{1}{2} \|(\mathbf{D}_n, C_n) - (\mathbf{D}'_n, C'_n)\|
\]
That is, the contraction constant is \(\frac{1}{2} < 1\).

Since the joint mapping is a contraction in the \((N+1)\)-dimensional space\cite{banach1922banach}, by the Banach fixed-point theorem, the sequence \((\mathbf{D}_n, C_n)\) converges to a unique fixed point \((\mathbf{D}^*, C^*)\), regardless of the initial values.

\section{Training Details}\label{othertrainingdeatils}
The prompt template used for our zero RL training is shown below:

\begin{tcolorbox}[
    colback=gray!10,
    colframe=gray,
    boxrule=0.5pt,
    boxsep=3.5pt,
    title=Training prompt of our zero RL,
    fonttitle=\bfseries,
    left=5pt, right=5pt,
    top=5pt, bottom=5pt
]
Question:\textbackslash n\{question\} 

Please reason step by step, and put your final answer within \textbackslash\textbackslash boxed\{\}.

Answer:\textbackslash n

\end{tcolorbox}

\Cref{tab:config} presents the key configuration used for our Qwen2.5-7B experiment. Training was conducted on a single node with 8 A100 GPUs, and the model was trained for 110 steps using the veRL library \cite{sheng2024hybridflowverl}.
For Qwen2.5-14B experiment, training was conducted on four nodes, each equipped with 8 A100 GPUs. The model was trained for 200 steps with a batch size of 256. Other configurations were kept consistent with those used in the Qwen2.5-7B experiment.

\begin{table*}[h]
\centering
% 自定义代码块样式
\lstset{
    basicstyle=\ttfamily\footnotesize, % 使用等宽字体和小号字体
    numbers=left, % 在左侧显示行号
    numberstyle=\tiny\color{gray}, % 行号使用灰色和更小的字体
    keywordstyle=\color{blue}, % 关键字使用蓝色
    commentstyle=\color{green!50!black}, % 注释使用暗绿色
    stringstyle=\color{red}, % 字符串使用红色
    backgroundcolor=\color{gray!10}, % 设置代码背景为浅灰色
    frame=tb, % 在代码块的顶部和底部添加边框
    rulecolor=\color{black}, % 边框颜色为黑色
    tabsize=4, % 制表符大小
    captionpos=b, % 标题位置在底部
    breaklines=true, % 自动换行
    breakatwhitespace=true, % 只在空格处换行
    showspaces=false, % 不显示空格
    showstringspaces=false, % 不显示字符串中的空格
    showtabs=false, % 不显示制表符
    morekeywords={*,...} % 如果有其他关键字需要高亮，可以在这里添加
}
\begin{lstlisting}[language=bash]
python -m verl.trainer.main_ppo \
    algorithm.adv_estimator=grpo \
    data.train_batch_size=1024 \
    data.val_batch_size=500 \
    data.max_prompt_length=1024 \
    data.max_response_length=4096 \
    actor_rollout_ref.actor.optim.lr=5e-7 \
    actor_rollout_ref.model.use_remove_padding=True \
    actor_rollout_ref.actor.ppo_mini_batch_size=256 \
    actor_rollout_ref.actor.ppo_micro_batch_size_per_gpu=4 \
    actor_rollout_ref.actor.use_kl_loss=True \
    actor_rollout_ref.actor.kl_loss_coef=0.001 \
    actor_rollout_ref.actor.entropy_coeff=0.001 \
    actor_rollout_ref.actor.clip_ratio=0.2 \
    actor_rollout_ref.actor.kl_loss_type=low_var_kl \
    actor_rollout_ref.model.enable_gradient_checkpointing=True \
    actor_rollout_ref.actor.fsdp_config.param_offload=False \
    actor_rollout_ref.actor.fsdp_config.grad_offload=False \
    actor_rollout_ref.actor.fsdp_config.optimizer_offload=False \
    actor_rollout_ref.rollout.temperature=1.0 \
    actor_rollout_ref.rollout.log_prob_micro_batch_size=160 \
    actor_rollout_ref.rollout.tensor_model_parallel_size=4 \
    actor_rollout_ref.rollout.enable_chunked_prefill=False \
    actor_rollout_ref.rollout.max_num_batched_tokens=5120 \
    actor_rollout_ref.rollout.name=vllm \
    actor_rollout_ref.rollout.gpu_memory_utilization=0.8 \
    actor_rollout_ref.rollout.n=8 \
    actor_rollout_ref.ref.log_prob_micro_batch_size=160 \
    actor_rollout_ref.ref.fsdp_config.param_offload=True \
    algorithm.kl_ctrl.kl_coef=0.001 \
    critic.ppo_micro_batch_size_per_gpu=4 \
    trainer.critic_warmup=0 \
    trainer.n_gpus_per_node=8 \
    trainer.nnodes=1 \
    trainer.remove_clip=False
\end{lstlisting}
\caption{Key configuration for our experiment.}
\label{tab:config}
\vspace{18pt}
\end{table*}

\section{Evaluation Benchmarks}\label{benckmarks}
We evaluate mathematical problem-solving ability using a variety of well-known benchmarks, ranging from middle school to Olympiad-level difficulty:

\begin{itemize}
    \item \textbf{AIME (American Invitational Mathematics Examination)}: A challenging high school level competition featuring 30 problems with integer answers. It is a well-established benchmark for evaluating advanced mathematical reasoning. Available at \url{https://huggingface.co/datasets/math-ai/aime25} and \url{https://huggingface.co/datasets/math-ai/aime24}.

    \item \textbf{AMC (American Mathematics Competitions)}: A benchmark focuses on mathematical problem-solving skills for middle and high school students. Problems range in difficulty and emphasize logical reasoning. Dataset available at \url{https://huggingface.co/datasets/math-ai/amc23}.

    \item \textbf{MATH500}~\cite{hendrycksmathdataset2021}: A curated subset of the larger MATH dataset, consisting of 500 problems that span various mathematical fields such as algebra, geometry, and number theory.

    \item \textbf{MinervaMath}~\cite{lewkowycz2022solvingminerva}: A benchmark used to evaluate the performance of large language models on detailed, multi-step quantitative reasoning problems.

    \item \textbf{OlympiadBench}~\cite{he2024olympiadbench}: A benchmark specifically designed for assessing models on Olympiad-level mathematics, including problems from national and international math competitions.

    \item \textbf{GSM8K}~\cite{cobbe2021traininggsm8k}: A dataset of grade school math word problems, commonly used to test arithmetic and step-by-step reasoning in models.
\end{itemize}

\section{Effectiveness of Symmetric Sampling}
\label{symmetric_sampling}

We further investigate the effectiveness of alignment-based symmetric sampling. For comparison, we directly select the \(|B|\) problems with the smallest value of \(A\) for the rollout phase. \par
\begin{figure}[htbp]
    \centering
    \includegraphics[width=0.55\linewidth]{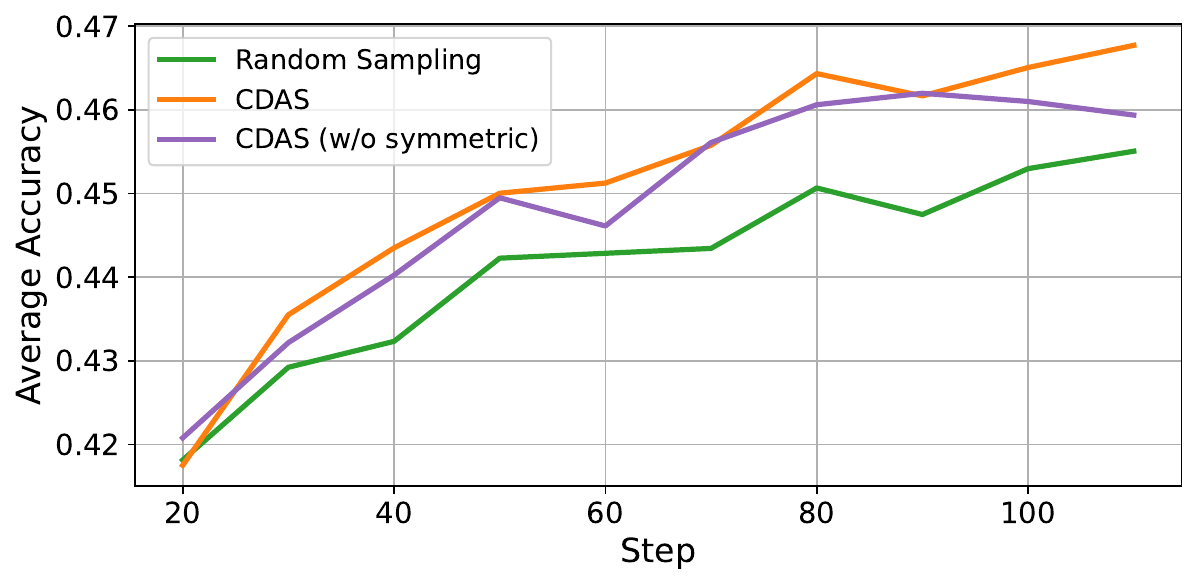}
    \caption{Ablation study of symmetric sampling.}
    \label{fig:ablation-symmetric}
\end{figure}

As seen in \cref{fig:ablation-symmetric}, after removing symmetric sampling, \ourmethod \ consistently outperforms the Random Sampling baseline. However, there is still an obvious decline in accuracy, especially in the later stages of training (after \(100\) steps). We attribute this phenomenon to the accumulated estimation errors in model competence due to the imbalanced sampling. In the early stages of training, imbalanced sampling can cause the model's competence to be either overestimated or underestimated. This discrepancy from the true competence accumulates as the number of training steps increases and hinders the convergence of \ourmethod, impairing the performance in the later stages. 

\section{Effectiveness of Warmup Phase}
\label{warmup_phase}
Our warm-up phase is specifically designed to address two potential criticisms. First, it directly mitigates the risk of inheriting and amplifying dataset biases. Without this initial calibration, the first few random batches could, by chance, overrepresent certain problem types (e.g., algebra in the MATH dataset). \ourmethod\ might then prematurely assign low difficulty to these common problems and high difficulty to unseen, underrepresented topics, effectively marginalizing them from future sampling. Our strategy prevents this by guaranteeing a comprehensive, initial assessment across the \textit{entire} problem distribution, ensuring that a problem's initial difficulty is based on an actual model attempt, not on an absence of data.
Second, this approach enhances the algorithm's generalizability and reduces its dependence on a dataset's specific difficulty distribution. The warm-up acts as a mandatory ``discovery phase," allowing \ourmethod\ to construct a complete and fair map of the initial difficulty landscape before it starts making targeted decisions. This bootstraps the system with a high-quality, holistic foundation, ensuring that the subsequent efficiency gains from competence-difficulty alignment are built upon a stable and equitable understanding of the task.

The importance of this well-founded initial state is not merely practical; it is also theoretically motivated. Since \ourmethod\ is built on a fixed-point system, the choice of initial values can affect its convergence speed and stability. To investigate the impact of this deliberate initialization, we conduct an ablation study by removing the warm-up phase and instead starting with naive zero-initialized difficulties.

\begin{figure}[htbp]
    \centering
    \subfigure{
        \includegraphics[width=0.4\textwidth]{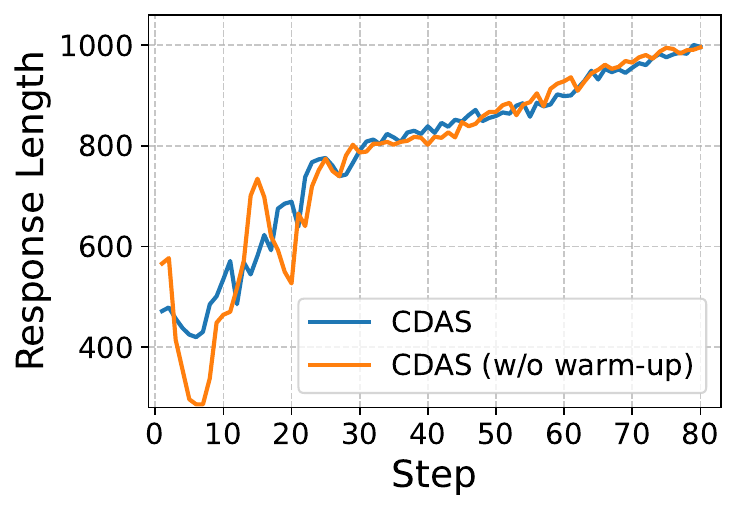}
        \label{fig:a}
    }
    \hspace{0.02\textwidth}
    \subfigure{
        \includegraphics[width=0.4\textwidth]{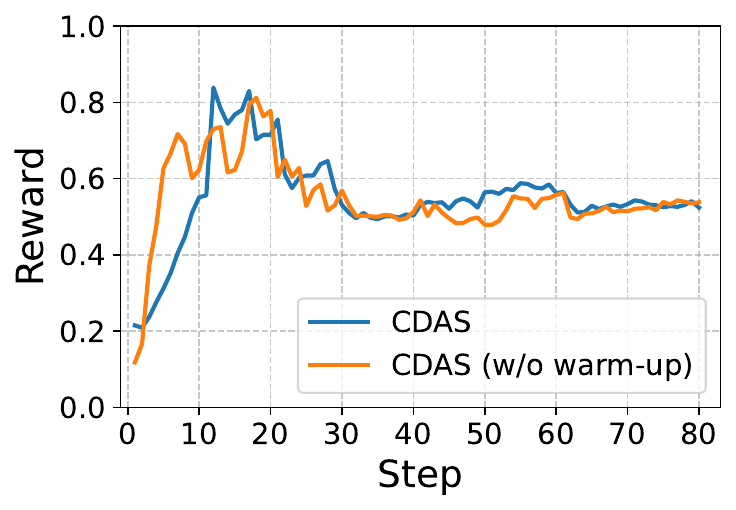}
        \label{fig:b}
    }
    \caption{Ablation study of the warm-up phase}
    \label{fig:ablation-warmup}
\end{figure}
As observed in \cref{fig:ablation-warmup}, the training curve exhibits significant fluctuations in the early stages of training (the first \(20\) steps) without the warmup stage. Specifically, in the first \(6\) steps, the removal of the warmup phase leads to a notable decrease in response length along with a rapid increase in reward, suggesting that the model has over-learned from simpler samples. However, as the convergence of the entire fixed-point system is guaranteed, the two training curves gradually overlap, further corroborating the stability of our framework.


\end{document}